RESEARCH ARTICLE

# Revolutionizing Wireless Networks with Federated Learning: A Comprehensive Review

Sajjad Emdadi Mahdimahalleh*

**ABSTRACT**

These days, with the rising computational capabilities of wireless user equipment such as smartphones, tablets, and vehicles, along with growing concerns about sharing private data, a novel machine learning model called federated learning (FL) has emerged. FL enables the separation of data acquisition and computation at the central unit, which is different from centralized learning that occurs in a data center. FL is typically used in a wireless edge network where communication resources are limited and unreliable. Bandwidth constraints necessitate scheduling only a subset of UEs for updates in each iteration, and because the wireless medium is shared, transmissions are susceptible to interference and are not assured. The article discusses the significance of Machine Learning in wireless communication and highlights Federated Learning (FL) as a novel approach that could play a vital role in future mobile networks, particularly 6G and beyond.

## 1. INTRODUCTION

Wireless communication is a technology that has had a significant impact on society over the past few decades. The rollout of the fifth generation (5G) of mobile technology, which promises reduced latency and higher data rates compared to previous generations, has already been done or is in process by many mobile operators. 5G and its beyond technology is expected to connect billions of heterogeneous devices to the network, supporting new applications and verticals under the Internet of Things (IoT) banner. This exponential growth in the number of connected IoT devices will enable new applications such as mobile healthcare, virtual and augmented reality, self-driving cars, smart buildings, factories, and infrastructures.

At the same time, modern machine learning (ML) techniques have led to significant breakthroughs in all areas of science and technology. New ML techniques continue to emerge, paving the way for new research directions and applications, including autonomous driving, finance, marketing, and healthcare. It is natural to expect that such a powerful tool as ML should also have a transformative impact on wireless communication systems, like other technology areas.

However, wireless communication system design has traditionally followed a model-based approach with great success. In communication systems, there is a good understanding of the channel and network models, which follow fundamental physical laws. Some of the current systems and solutions already approach fundamental information theoretical limits. Additionally, communication devices follow a highly standardized set of rules. Standardization dictates what type of signals to transmit and when and how to transmit them, and requires tight coordination across devices and even different networks. As a result, existing solutions are products of research and engineering design efforts that have been optimized over many decades and generations of standards based on theoretical foundations and extensive experiments and measurements.

From this perspective, compared to other areas in which ML has made significant advances in recent years, communication systems can be considered more amenable to model-based solutions rather than generic ML approaches. As a result, one can question the potential benefits ML can provide to wireless communications [1]–[3].

First, despite advances in coding and communication techniques, current design approaches in complex networks of heterogeneous devices still fall short of approaching theoretical limits. The modular design used in point-to-point communication systems is suboptimal, as it relies on separate optimization of individual blocks. With 5G and 6G, which are aiming to connect billions of IoT devices, a more adaptive and efficient use of resources is necessary. Cross-layer design is an alternative, but it often leads to ad-hoc solutions. ML techniques can provide a data-driven solution with potentially larger performance gains and reasonable implementation complexity.

Second, current communication systems often struggle to find optimal solutions even with a modular approach. Theoretical solutions for problems such as maximum likelihood detection in multiuser scenarios and optimal beamforming vectors in multi-antenna transmitter networks exist but are computationally prohibitive [4], [5]. Consequently, solutions to these problems are often low-complexity and suboptimal [6] or simple closed-form solutions [7]. Recent research suggests that machine learning techniques can offer more favorable trade-offs between complexity and performance [8], [9].

However, machine learning (ML) approaches can offer another potential benefit by going beyond the modular design and learning the optimal communication scheme in an end-to-end fashion rather than targeting each module separately [10], [11]. This end-to-end ML approach enables the tackling of much more complex problems with solutions that were previously considered unattainable with traditional approaches. For instance, the joint source-channel coding problem, which is typically divided into the compression and channel coding subproblems, has been a challenging task for decades. Despite efforts to design joint source-channel coding schemes, they often resulted in ad-hoc solutions for different source and channel combinations with significant complexity. Recent results have shown that ML can provide significant improvements in the end-to-end performance of joint source-channel coding [12]–[15].

ML has the potential to compensate for unknown and varying channel knowledge without the need for knowledge of channel parameters or statistics, which is required in many traditional communication methods. This is particularly relevant in modern wireless networks where hardware limitations can make it difficult to design a communication system based on exact channel knowledge. Additionally, there is potential for synergy between communication networks and ML, where communication networks can be designed to enable ML applications. Currently, most of the learning in mobile devices is carried out centrally at cloud servers, which can be problematic due to the sheer volume of collected data. To address this issue, it is necessary to enable edge devices to extract and convey only the relevant information for the underlying learning task to reduce the amount of data that needs to be offloaded to the cloud.

A major issue with centralizing intelligence is the potential for privacy and security breaches, as sensitive personal information is collected from edge devices. This can discourage consumers from using these services. Local learning at individual devices is an alternative, but limited data may lead to overfitting. Centralized intelligence also causes latency, which can be problematic in applications that require rapid inference and action. To address these issues, the objective of merging communications and ML is to bring intelligence to the network edge, where massive datasets and processing power are available in a distributed manner.

## 2. Overview of Machin Learning

Artificial intelligence (AI) refers to techniques that enable computers to perform tasks that typically require human intelligence, such as reasoning, problem-solving, and learning. Machine learning is a subset of AI that focuses on algorithms that can learn from data without being explicitly programmed. Deep learning, a type of machine learning, involves training neural networks to extract patterns from data.

In wireless communications, machine learning algorithms have been used to address a wide range of problems. These problems can be categorized into four main types: supervised learning, unsupervised learning, semi-supervised learning, and reinforcement learning (RL).

Overall, machine learning has shown great promise in improving wireless communication systems, and each category of learning algorithms offers unique benefits and applications. In the following sections, each category will be divided deeper.

### 2.1. Supervised Learning

Supervised learning involves training an algorithm to learn the relationship between input and output for a function, which can apply to various types of input and output. In this type of problem, the model is trained on labeled data where the input and output pairs are provided. For instance, an input vector $x \in X$ and a corresponding vector of target variables $c \in C$ can be considered. Supervised learning tasks are divided into two groups: classification and regression. If the input is mapped to a finite number of discrete classes, then it's a classification task (C is a discrete set). If the output can have continuous values, then it's a regression task (C is a continuous set). An example of a classification task is the problem of classifying images into categories such as dog or cat. Common examples for benchmarking supervised learning algorithms include the classification of handwritten digits and spam detection.

In communications, a receiver designed for a fixed transmission scheme can be seen as a classification problem. For instance, in a simple modulation scheme where input bits are mapped to constellation points, the receiver needs to classify each received noisy symbol to one of the constellation points [1]. This classification task can also be extended to decoding of coded messages sent over a noisy channel, where the decoder function attempts to map a vector of received symbols to a codeword [16], [17].

Wireless communication systems often encounter regression problems, such as channel estimation [10], [11], [18], which involves estimating channel coefficients from noisy received versions of known pilot signals. Traditional methods assume a known channel model and estimate its parameters using least squared or minimum mean-squared error techniques. However, data-driven channel estimation methods do not rely on any assumptions about the channel model and instead use training data from the underlying channel. Although this approach is useful when an accurate channel model is unavailable, it requires a large training dataset and may need to be conducted off-line.

Supervised learning can offer a useful approach to quickly generate approximate suboptimal solutions for complex optimization problems. In wireless networks, many optimization problems are highly complex and non-convex or combinatorial, such as scheduling or determining transmit powers in a network of interfering transmitters. These problems often do not have low-complexity optimal solutions, but in practice, we need a fast solution to be implemented within the time constraints of the channels. Typically, a low-complexity suboptimal solution is used, but an alternative approach is to train a neural network using the optimal solution as supervision. This can yield a reasonable performance that can be quickly obtained once the network is trained. Such methods have been extensively applied to wireless network optimization problems, with promising results in terms of the performance-complexity trade-off [8], [19].

### 2.2. Unsupervised Learning

In this type of problem, the model is trained on unlabeled data, and the objective is to find patterns or structures in the data without any prior knowledge of the output. In other words, unsupervised learning involves training data without any output values, and the goal is to learn functions that describe the input data. These functions may be useful for more efficient supervised learning and for making the input data more amenable to human understanding. Unsupervised learning is often used for clustering, dimension reduction, and density estimation, which have all been used extensively in communication systems. Clustering is essentially source compression, where the goal is to identify a small number of representatives that can adequately represent all possible input vectors. Density estimation aims to determine the distribution of data as accurately as possible from a limited number of samples, while dimensionality reduction aims to limit the dimension of a representation of data. Autoencoders are a common neural network approach to dimensionality reduction, which plays an important role in the data-driven design of compression and communication schemes. Generative models learn the distribution of data and have been used with deep learning in deep generative models, which have shown remarkable performance in modeling complex distributions.

### 2.3. Semi-Supervised Learning

This type of problem lies between supervised and unsupervised learning. In this type of problem, the model is trained on partially labeled data, where only some input and output pairs are provided. The objective is to use the labeled data to improve the model's performance on unlabeled data. This type of learning is commonly used in natural language processing. In speech recognition, semi-supervised learning has been used for speaker identification and language modeling.

Semi-supervised learning can be particularly useful when obtaining labeled data is difficult, expensive, or time-consuming. In some cases, it may be easier to collect a large amount of unlabeled data than to obtain labeled data. Semi-supervised learning can help leverage this data by using a small amount of labeled data to guide the learning process for the larger pool of unlabeled data.

This approach can also be helpful when the labeled data may not fully represent the diversity of the target population or when the labeled data is noisy or of low quality. By leveraging unlabeled data, semi-supervised learning can help improve the model's robustness and generalization ability.

Overall, semi-supervised learning is a powerful technique that can help improve the performance of machine learning models when labeled data is scarce or difficult to obtain.

### 2.4. Reinforcement Learning (RL)

RL is a type of machine learning where the goal is to learn how to interact in a random, unknown environment based on feedback received in the form of costs or rewards following each action. The agent interacts with the environment by taking actions, and the goal is to learn the right action to take at each state in order to maximize (minimize) the long-term reward (cost). RL has found applications in wireless networks as early as in the 1990s [20], [21], including power optimization in the physical layer for energy-efficient operation [22]. RL algorithms can be used for two types of problems requiring interactions with an environment. In one type, the device might have an accurate model of the environment, but the solution of an optimal operation policy for the device may be elusive. In such a case, RL techniques can be used as a numerical solution technique to characterize the optimal (or near-optimal) strategy for the device. This type of problem typically appears in networking, where multiple devices are scheduled to share limited spectral resources.

## 3. Challenges of Traditional Centralized Machine Learning in Wireless Networks

Over one billion devices are now connected to the internet, and as the use of mobile devices increases, the amount of data transmitted wirelessly is growing exponentially. This data is generated by the Internet of Things (IoT) paradigm, which includes applications such as wearable technologies, factory automation, and autonomous systems. While much of the current data traffic involves video and voice content for user consumption, IoT-generated data is intended for machine analysis and inference to enhance IoT applications. Currently, a common approach to implement IoT intelligence is to transmit all relevant data to a cloud server and use available processing power to train a powerful machine learning (ML) model. However, due to communication latency and the limited power and bandwidth of IoT devices, a centralized solution is not always applicable, particularly as the volume of collected data increases. Additionally, centralized approaches raise privacy concerns, as users may be hesitant to share their data with third parties. To address these challenges, a promising approach is to bring intelligence to the network edge by enabling wireless devices to implement distributed and collaborative learning algorithms [1], [2]. The network edge is rich in both massive datasets and computational power, but these resources are highly distributed, making the development of seamless and efficient learning algorithms a critical challenge for edge intelligence.

Developments in distributed optimization algorithms have provided the ability to train joint ML models collaboratively while keeping data local to each participating device, thus partially addressing privacy concerns. These algorithms also allow for computational scalability by leveraging computational resources across many edge devices. However, the overall learning speed does not increase linearly with the number of devices due to communication bottlenecks that can limit computation speed. Moreover, wireless devices are highly heterogeneous in terms of their computational abilities, and their computation speed and communication rates can be highly variable due to physical factors. As a result, distributed learning algorithms designed for wireless network edge implementation must be thoughtfully designed to account for the impact of the variable communication network and the stochastic computation capabilities of devices.

### 3.1. Collaborative Learning

Stochastic optimization problems are commonly used to model parameterized machine learning (ML) problems. These problems seek to minimize a stochastic loss function, $F$, over a set of model parameters, $\theta$, using random data samples, $\zeta$, drawn from an unknown distribution, $P$. In practice, we often only have access to a dataset, $D$, sampled from $P$, and instead minimize the empirical loss function, $E_{\zeta \sim D} F(\theta, \zeta)$, using stochastic gradient descent (SGD).

In a distributed setting with $N$ devices denoted by set $S$, each with its own local dataset, $D_i$, the problem can be solved in a distributed manner by redefining the minimization problem as an average of $N$ stochastic functions:

$$\min_{\theta \in R^d} f(\theta) = \frac{1}{N} \underbrace{\sum_{i=1}^{N} E_{\zeta \sim D_i} F(\theta, \zeta)}_{:= f_i(\theta)} \quad (1)$$

The goal of collaborative learning is to solve this minimization problem in a distributed manner, where each device minimizes its own local loss function, $f_i(\theta)$, while seeking a consensus with other devices on the global model, $\theta$. Consensus can be achieved through a central entity such as a parameter server (PS), or through communication with a limited number of neighboring devices. Collaborative learning methods can be divided into two classes based on the consensus framework, which is called centralized and decentralized. This review will provide a brief overview of centralized learning, followed by a more in-depth discussion of federated learning as a centralized distributed learning strategy.

Before delving into the details of centralized learning, it is important to understand the concept of decentralized learning. In a decentralized learning scenario, the main difference lies in the consensus step, where each device combines its local models with those of its neighbors based on a certain rule determined by the network topology. This approach is different from centralized learning, where a star topology is typically used to fully synchronize the models across devices after each global averaging step. In decentralized learning, the lack of full synchronization introduces additional noise in the framework.

### 3.2. Centralized Learning

In distributed learning, a framework is considered centralized if a PS manages the learning. In this approach, each user performs local computation under the PS's supervision, and the results are then utilized by the PS to solve the optimization problem. One of the most widely used centralized strategies for training DNN architectures is parallel synchronous stochastic gradient descent (PSSGD).

In PSSGD, at the start of iteration t, each device retrieves the current global model $\theta_t$ from the PS and computes its local gradient estimate to minimize its own loss function $f_i(\theta_t)$.

$$g_{i,t} = \nabla_\theta F(\theta_t, \zeta_{i,t}) \quad (2)$$

After sampling $\zeta_{i,t}$, each device computes a local gradient estimate for its loss function using the global model $\theta_t$. The local gradients are then sent to the PS to be aggregated, and the resulting gradient is used to update the global model $\theta_t$. It is worth noting that the devices can use mini batches with multiple samples to compute the local gradient estimate.

$$\theta_{t+1} = \theta_t - \eta_t \cdot \frac{1}{|S|} \sum_{i \in S} g_{i,t} \quad (3)$$

where $\eta_t$ is the learning rate at iteration t.

## 4. Federated Learning

FL is a centralized distributed learning strategy that solves the stochastic optimization problem in a distributed manner without sharing local datasets to ensure user privacy, which is especially relevant for sensitive data like medical records used for healthcare or remote diagnosis applications [23], [24]. FL differs from other centralized learning strategies as it practices collaborative learning on a large scale, necessitating additional mechanisms to make it communication efficient. Local SGD strategy is used to reduce communication frequency, while to prevent PS overload, only a subset of active devices in the network, denoted by $S_t \subseteq S$, participate in the learning process at each iteration $t$. Device selection can be random or more advanced based on the problem that is going to be solved.

In general, FL utilizes distributed SGD, and each iteration of SGD comprises three main steps: device selection or sampling, local computation, and aggregation or consensus. After the PS selects the devices $S_t \subseteq S$ for the $t$th iteration, each device first retrieves the most recent global model $\theta_{t-1}$ from the PS, and then performs local SGD update to minimize its own loss function $f_n(\theta_t)$:

$$\theta_{n,t}^h = \theta_{n,t}^{h-1} - \eta_t \cdot g_{n,t}^h \quad (4)$$

for $h = 1, \ldots, H$, where we set $\theta_{i,t}^0 = \theta_{t-1}$:

$$g_{n,t}^h = \nabla_\theta F(\theta_{n,t}^h, \zeta_{n,h}) \quad (5)$$

and $\zeta_{n,h}$ is the data sampled from dataset $D_n$ in the $h$th local iteration.

After each device completes H SGD steps in an iteration, it sends its latest local model $\theta_{n,t}^h$ to the PS for aggregation. The aggregation or consensus in FL is commonly performed by taking the average of the participating devices' local models:

$$\theta_{t-1} = \frac{1}{|S_t|} \sum_{i \in S_t} \theta_{i,t}^H \qquad (6)$$

The generic FL framework iterating over these three steps is called federated averaging (FedAVG).

### 4.1. Device Selection and Resource Allocation in Distributed Learning Over Wireless Networks

The optimal balance between the accuracy of the model update and communication cost is a crucial design problem in centralized collaborative learning. It involves determining the optimal number of uplink devices and the frequency of global model updates, which impacts the convergence performance based on wall clock time. When collaborative learning is performed across wireless devices, the communication bottleneck becomes even more restrictive due to limited bandwidth, varying channel quality, and energy limitations of most wireless devices. Hence, designing optimal collaborative learning across wireless networks requires optimizing wireless networking parameters along with collaborative learning algorithm parameters, considering the channel conditions of the devices [25]–[28]. While random and round-robin scheduling are the most commonly used selection mechanisms of devices in FL, they do not consider the channel condition of individual devices. The synchronization required for the global model update in each iteration can result in significant delays when devices with weaker channel conditions become the bottleneck.

To address the issue of devices with weaker channel conditions becoming the bottleneck in FL, the device selection mechanism needs to be modified to consider the channel conditions of each device.

In FL, a relationship between the number of communication rounds and the number of scheduled devices can be established to attain the desired level of final accuracy. However, this assumes that devices are scheduled independently and with equal probability. This ensures that devices participate equally in the long run to avoid any bias in the overall process, even if one device is preferred over another at a certain iteration. Nevertheless, in wireless settings, scheduling statistics may not be identical due to heterogeneity, such as differences in the distance of devices from the access point acting as the PS. Thus, efficient device selection strategies for FL must consider communication constraints and account for such heterogeneity.

### 4.2. Scheduling Policies for FL in Wireless Networks

To enable FL, new challenges need to be addressed that differ from traditional distributed optimization methods [29]. FL involves training on a large, non-i.i.d. and often unbalanced dataset generated by different distributions across user equipment (UEs). Additionally, the AP needs to link a vast number of UEs through a resource-constrained spectrum and can only allow a limited number of UEs to send their trained weights via unreliable channels for global aggregation. These challenges make issues such as stragglers and fault tolerance significantly more important than for conventional training in data centers. Considerable research has been carried out to address these challenges, mainly in two directions: algorithmic and communication. Algorithmic methods range from reducing communication bandwidth by updating only UEs with significant training improvement [30] to compressing gradient vectors via quantization [31] or adopting a momentum method in the sparse update to accelerate training [32]. Communication methods include adapting the number of locally computing steps to the variance of the global gradient [33]–[35] or scheduling the maximum number of UEs in a given time frame [36]. New methods have also been developed that exploit compute-over-air mechanisms and a jointly decode-and-average scheme at the edge computing unit when spectral resources become the communication bottleneck [37], [38]. To fully realize the potential of federated learning, it is necessary to scale up the deployment across an extensive decentralized network, where communications are subject to inter-cell interference and can encounter failure due to the shared nature of the wireless medium. Therefore, a complete understanding of the performance of FL when operating under different scheduling schemes with unreliable communication links is essential for its successful delivery in large-scale wireless networks [39].

### 4.3. Scheduling Policies

In real-world systems, communication between machines is significantly slower than local computing [27]. Therefore, updating trained parameters from all UEs sequentially can cause significant overhead in communication time. Instead, the AP should select only a subset of UEs and update their parameters simultaneously to keep communication time within acceptable limits. To achieve this, scheduling policies play a crucial role in assigning resource-limited radio channels to appropriate UEs. Three practical scheduling criteria, denoted by G = K/N (K is number of associated UEs per Access Point (AP), N number of subchannels, and G is the UE number over subchannel number ratio), are considered: Random Scheduling (RS), Round Robin (RR), and Proportional Fair (PF) [40], [41]. In RS, the AP randomly selects N associated UEs in each communication round, assigns each selected UE a dedicated subchannel for parameter transmission. In RR, the AP divides all UEs into G groups and consecutively assigns each group to access the radio channels and update their parameters per communication round. In PF, the AP selects N out of the K associated UEs according to followed policy during each communication round [61].

$$m^* = \underset{m \subset \{1,2,...,K\}}{\operatorname{argmax}} \left\{ \frac{\tilde{\rho}_{m_1,t}}{\overline{\rho}_{m_1,t}}, \ldots, \frac{\tilde{\rho}_{m_N,t}}{\overline{\rho}_{m_N,t}} \right\} \qquad (7)$$

where $m = (m_1, \ldots, m_N)$ is a length-$N$ vector and $m^* = (m_1^*, \ldots, m_N^*)$ represents the indices of the selected UEs,

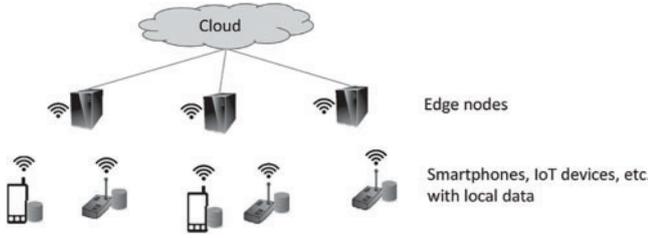

Fig. 1. Model training at the edge.

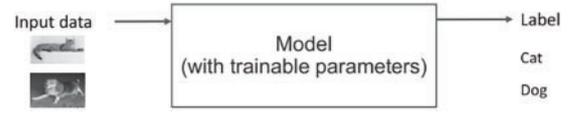

Fig. 2. Machine learning model.

$\tilde{\rho}_{m_i,t}$ and $\overline{\rho}_{m_i,t}$ are the instantaneous and time average signal-to-noise ratio (SNR) of UE $m_i$ at the communication round $t$, respectively [41].

Overall, the choice of scheduling strategy will depend on the specific requirements of the federated learning problem, as well as the available computational resources and communication bandwidth [61].

As illustrated in Fig. 1, a 5G and beyond system with mobile edge computing (MEC), edge nodes located at base stations or elsewhere in close proximity to users act as processing units with a high-bandwidth and low-latency connection to end devices [61].

### 4.4. Model Training at the Edge

After gaining a basic understanding of learning and distributed learning, as well as the concept of federated learning as a novel learning technique in mobile edge systems and scheduling policies, the focus will now shift towards exploring federated learning algorithms.

### 4.5. Federated Learning

Federated learning (FL) is a challenging and dynamic task in wireless edge environments. The need for multiple rounds of computation and communication between clients and the server adds complexity to the process. The distributed clients, known as end devices or edge nodes, generate or collect data and process it locally before sending the learned model parameters through wireless links to the server for aggregation. This process is resource-intensive and requires careful management of communication, computation, and energy resources. As a result, the quality of the FL process is influenced by the trade-offs among these resources.

To address this challenge, researchers have explored adaptive federated learning techniques. These techniques aim to optimize the performance of FL by dynamically adapting to changes in the network and resource availability. Adaptive FL algorithms adjust the number of clients involved in each round, the communication schedule, and the local training process to improve the overall performance of the system. In the next sections, a mathematical description of FL is provided.

### 4.6. Mathematical Definition of Federated Learning

Typically, machine learning (ML) systems consist of models that can be trained to map input data to the desired output, as shown in Fig. 2. Each model has its own computational logic and a set of parameters that can be adjusted during the training process. In this process, the model is trained using a dataset that contains input data samples and their corresponding ground-truth labels. Once trained, the model can be used to predict labels for new data samples with unknown labels.

Let $g(w, x)$ represent the model with a trainable parameter vector w, which maps the input data sample x to the predicted output label $\hat{y}$, as shown in Fig. 2. The per-sample loss is defined as $l(\hat{y}, y) = l(g(w, x), y)$, where y is the ground-truth output (label) available in the training data but not during the inference phase. Common error functions like mean square error, cross-entropy, etc., can be used [43], and additional terms related to model parameter w, like regularization, can be included. Hence, it is more convenient to express the loss function as $l(w, x, y)$ instead of relating them through $g(\cdot, \cdot)$. Table I provides examples of per-sample loss functions $l(w, x, y)$ of some popular ML models.

As the training dataset has multiple samples, researchers are interested in the average of per-sample losses of all data samples in the training dataset. In the federated learning setting, assume that there are $N$ clients in total, where each client $i \in \{1, 2, \ldots, N\}$ has its local training dataset $S_i$, including data samples $\{x, y\} \in S_i$ that capture both the input $x$ and its desired output/label $y$.

Based on this, the local average loss of client $i$ can be defined as

$$f_i(w) := \frac{1}{|S_i|} \sum_{\{x,y\} \in S_i} l(w, x, y) \qquad (8)$$

where $|.|$ denotes the cardinality (i.e., the number of elements) in the set. It can further be defined the global average loss as

$$f(w) := \sum_{i=1}^{N} p_i f_i(w) \qquad (9)$$

where $p_i$ is a weighting coefficient, such that $\sum_{i=1}^{N} p_i = 1$. Assumption is that $S_i \cap S_j = \varnothing$ for $i \neq j$ and a few possible ways of choosing $p_i$ are discussed.

- If it is chosen $p_i = \dfrac{|S_i|}{\left|\bigcup_{j=1}^{N} S_j\right|}$, then:

$$f(w) = \frac{|S_i|}{\left|\bigcup_{j=1}^{N} S_j\right|} \sum_{\{x,y\} \in \bigcup_{j=1}^{N} S_j} l(w, x, y) \qquad (10)$$

Based on (9) and (10), the average loss is calculated when all local datasets (i.e., $\bigcup_{j=1}^{N} S_j$) are available in a central location. By defining $p_i$ according to this equation, all data samples are given equal importance, meaning that clients with fewer data samples have less importance during the FL process.

- If we set $p_i = 1/N$, all clients are equally important, regardless of the number of data samples they have. For example, a client with only one data sample is as important as a client with 100 data samples. This means that the data sample from the former client is more important than each data sample in the latter client.

TABLE I: PER-SAMPLE LOSS FUNCTIONS OF EXEMPLAR MODELS

| Model | Per-sample loss function $l(w, x, y)$ |
|---|---|
| Linear regression | $\frac{1}{2}\|y - \mathbf{w}^T \mathbf{x}\|^2$ |
| SVM (smooth loss) | $\frac{\lambda}{2}\|\mathbf{w}\|^2 + \frac{1}{2} max\{0; 1 - y\mathbf{w}^T \mathbf{x}\}^2$ ($\lambda \geq 0$ is a constant) |
| K-means (Q clusters) | $\frac{1}{2} min_q \|\mathbf{x} - \mathbf{w}_{(q)}\|^2$ where $:= [\mathbf{w}_{(1)}^T, \ldots, \mathbf{w}_{(q)}^T, \ldots, \mathbf{w}_{(Q)}^T]^T$ |
| Convolutional neural network | Cross-entropy on neural network output |

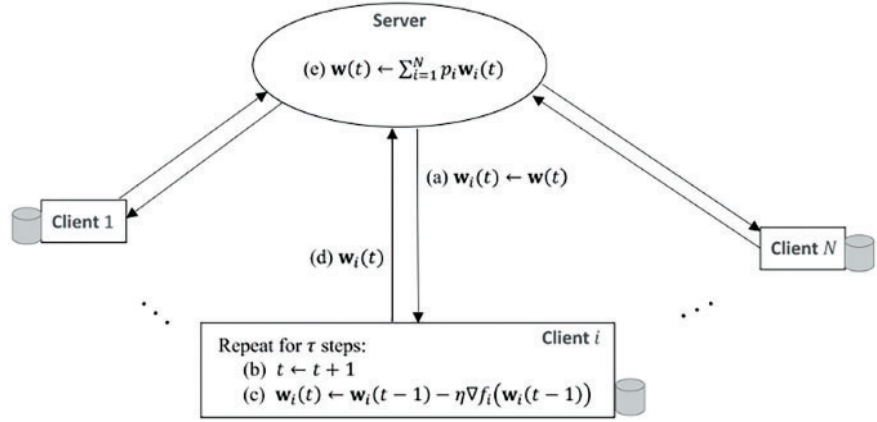

Fig. 3. Federated averaging (FedAvg).

- Choosing $p_i$ between $\frac{|S_i|}{\left|\bigcup_{j=1}^{N} S_j\right|}$ and $1/N$ balances the two cases.

Practitioners define $p_i$, and its choice controls the trade-off between sample bias and client bias.

### 4.7. Learning Problem

The FL problem can be formulated as finding the optimal parameter vector $w^*$ that minimizes the global loss $f(w)$, which is expressed as (11):

$$w^* := \underset{w}{\mathrm{argmin}}\, f(w) \qquad (11)$$

It is worth noting that both ML and FL problems are optimization problems. However, solving (10) poses a twofold challenge. Firstly, the complexity of ML models and data-dependent loss functions makes it difficult to obtain a closed-form solution. Secondly, in FL, the global loss function $f(w)$ is not directly observed as it is defined on distributed datasets at clients. Instead, only the local loss function $f_i(w)$ is observable at each client i. To tackle these challenges, gradient descent approaches are commonly used for ML problems [43]. In FL, a distributed version of gradient descent is often employed, so that the data is kept at clients and not shared with the central server.

### 4.8. Federated Averaging (FedAvg)

A common approach to solving (10) in the FL scenario is through the federated averaging (FedAvg) algorithm [44]. FedAvg consists of multiple rounds, with each round consisting of local iterations at each client followed by a parameter aggregation step that involves both clients and the server. The steps of FedAvg are depicted in Fig. 3 [45].

At the beginning of each round, the server sends its current parameter vector w(t) to all clients (Step (a) in Fig. 3), where $t$ is the iteration index that starts at $t = 0$ for the parameter vector during initialization. After receiving $w(t)$ from the server, each client $i$ sets its local parameter vector $w_i(t)$ to $w(t)$. Then, each client $i$ performs $\tau$ steps of gradient descent using its local loss function, defined on its local dataset, as shown in (8). Each gradient descent step updates the local parameter vector $w_i$ in accordance with (12):

$$w_i(t) = w_i(t-1) - \eta \nabla f_i(w_i(t-1)) \qquad (12)$$

The parameter vector from each client is updated using the gradient descent step size $\eta$ during $\tau$ iterations (as shown in Steps (b) and (c) in Fig. 3). The resulting parameter vector from each client is then sent to the server, where the server aggregates them to obtain the updated global parameter vector and depicted in steps (d) and (e) of Fig. 3:

$$w(t) = \sum_{i=1}^{N} P_i w_i(t) \qquad (13)$$

This iterative process continues until a stopping condition is met, which can be defined by a maximum number of rounds, a given resource budget, or when the decrease in the loss function is below a certain threshold. Let assume that the training stops after K rounds. For any $\tau$, there will be $T = K\tau$ local iterations (counted on any single client) before the stopping condition is met. The final model parameter can be either $w(T)$ obtained after aggregation in the last round or:

$$w^f = \underset{k \in \{1,2,\ldots,K\}}{\mathrm{argmin}}\, f(w(k\tau)) \qquad (14)$$

The parameter that yields the smallest global loss among all rounds is commonly used in practice, while the latter is valuable for certain types of theoretical analysis that we will discuss later. It's worth noting that computing $f(w(k\tau))$ is feasible if each client $i$ sends its local loss $f_i(w(k\tau))$ to the server. The rationale for this approach is that when $\tau = 1$ (i.e., parameter aggregation is performed after every local iteration), FedAvg is equal to gradient descent in the centralized setting. This is because the aggregation of (12) is equivalent to performing gradient descent on the global loss $f(\cdot)$, as gradient computation is linear. This equivalence, however, does not hold when $\tau > 1$. Thus, optimizing $\tau$ is necessary to achieve a good trade-off between communication overhead and learning convergence (see (9) and (13)).

It should be noted that standard FedAvg involves exchanging the entire parameter vector between the clients and the server at the end of each round. However, this process can be modified by exchanging a sparse gradient after every local iteration (i.e., $\tau = 1$) [17] or by exchanging a compressed model obtained by pruning the original model [42]. When $\tau > 1$ the gradient computed in a single iteration does not represent the change in the parameter vector across all iterations within the same federated learning round, the parameter vector or its difference between the start and end of the round needs to be communicated instead of just the gradient. In the case of large training datasets, such as in deep learning, stochastic gradient descent (SGD) is often used instead of deterministic gradient descent (DGD) [44] in (11).

### 4.9. Other Model Fusion Algorithms

In addition to FedAvg, there have been other model fusion techniques developed recently. For instance, FedProx enhances FedAvg to better handle FL with varied data and devices by introducing a proximal term to the objective function [46]. A Bayesian approach for efficient model fusion in FL settings has also been proposed [47]. Generally, these extensions build upon the fundamental FedAvg algorithm or at least its key concept of alternating between local computation and model fusion in every round.

The development of other model fusion algorithms for Federated Learning is an active area of research. These algorithms aim to improve the performance of FL by addressing specific challenges, such as heterogeneous data and device distributions. The success of these algorithms depends on their ability to strike a balance between the communication overhead, learning convergence, and other factors such as model complexity and data privacy. Therefore, it is essential to continue developing and exploring new techniques to advance the state-of-the-art in Federated Learning. Additionally, the development of new algorithms can lead to exciting new applications of FL in various domains, including healthcare, finance, and the Internet of Things (IoT).

### 4.10. Resource Constrained FL

When operating within a specified resource budget, a common query arises regarding the optimal approach to executing FL in order to minimize the global loss function while also adhering to the resource limitations. To strike the ideal balance between communication and computational resource consumption, a viable solution is to regulate the number of local iterations per ($\tau$).

Consider $M$ varieties of resources, which could represent time, energy, communication bandwidth, monetary expenses, or combinations of them. Each local iteration among clients involves consuming an amount $c_m$ the $m^{th}$ resource type, and each parameter aggregation step consumes an amount $a_m$ of the same resource type. A budget of $B_m$ is assigned to each resource type $m$. Based on these constraints, we can define the resource-constrained federated learning problem.

$$\min_{\tau, K \in \{1,2,3,...\}} f(w^f) \quad (15)$$

$(k\tau + 1)c_m + (k+1)a_m \leq B_m, \forall m \in \{1, \ldots, M\}$

The main objective of (15) is to minimize the global loss of the outcome obtained through FedAvg by optimizing $\tau$ and K.

However, determining the relationship between the objective function and $\tau$ and $K$ is challenging because it is dependent on how FedAvg converges over time. Typically, only upper bounds can be obtained for this convergence. Additionally, the analysis needs to consider the impact of different $\tau$ and $K$, which makes it more challenging than analyzing standard gradient descent approaches.

In FL systems, non-identical distribution of data across clients is another significant challenge, which is not an issue in distributed ML in data centers. This non-i.i.d distribution of data renders many convergence results for distributed ML invalid since they are derived based on independent and identically distributed (i.i.d) data. So the first challenge would be identifying a convergence bound with non-i.i.d data and a given $\tau$, but as this is just a review report, I just provide the main results.

In cases where the resource budget is infinite, selecting $\tau = 1$ is always advantageous. However, when dealing with finite resource budgets, it may be more beneficial to opt for a larger $\tau$ value since it can result in a smaller $f(w^f)$ within the constraints of the resource budget, stopping at some finite T. Therefore, for $\tau > 1$, the optimality gap increases with the gradient divergence, implying that $\tau$ should be smaller when the client data distribution is more non-i.i.d., leading to a larger gradient divergence. This is reasonable since the greater the diversity between clients, the faster their parameters diverge, and hence they require more frequent aggregation. These findings are also consistent with the experimental results that aims to identify the best $\tau$.

Fig. 4 illustrates the optimized FL approach with adaptive $\tau$, which combines the FedAvg process, parameter estimation in $G(\tau)$, resource consumption monitoring, and compliance with a resource budget. $G(\tau)$ is an optimization problem that will result to obtain $\tau$ based on the budget and K and also other parameters which its details are ignored in this review. The resource budget is considered an input to the system and can be defined by a system administrator or a separate job scheduling mechanism.

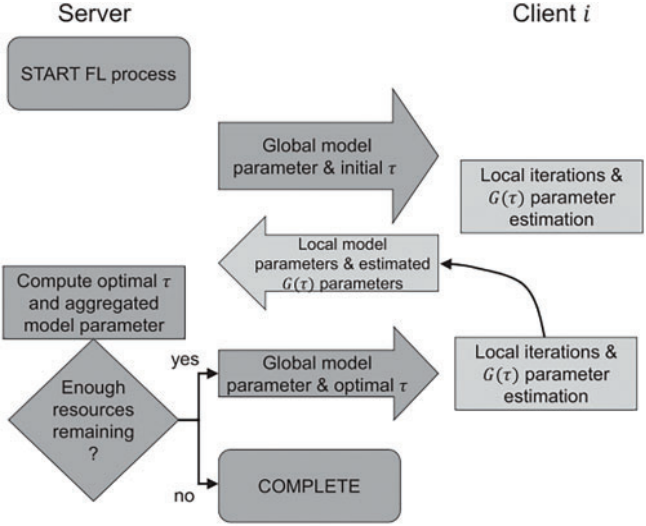

Fig. 4. Protocol of FedAvg with adaptive $\tau$.

Following a brief study and familiarization with federated learning, the subsequent section aims to present a case study that employs FL in wireless networks to address a specific problem, along with its corresponding outcomes.

## 5. FL for Channel Estimation in Conventional and RIS-Assisted Massive MIMO

The accuracy of the instantaneous channel state information (CSI) is critical in both conventional and RIS-assisted massive MIMO scenarios due to the highly dynamic nature of the mm-wave channel [48]. Reliable channel estimation accuracy is necessary for designing the analog and digital beamformers in conventional massive MIMO [49], [50], and the design of reflecting beamformer phase shifts of the RIS elements in the RIS-assisted scenario [51], [52]. RIS-assisted massive MIMO is more challenging since it involves signal reception through multiple channels (e.g., BS-RIS, RIS-user, and BS-user), which requires more advanced channel estimation schemes such as compressed sensing [52], angle-domain processing [53], and coordinated pilot-assignment [54]. Data-driven techniques, such as machine learning (ML) based approaches, have been proposed to provide robustness against imperfections or corruptions in the array data. These techniques can uncover the non-linear relationships in data/signals with lower computational complexity and achieve better performance for parameter inference [51], [52]. Compared to model-based techniques that heavily rely on mathematical models, ML has the following advantages:

1. It constructs a non-linear mapping between raw input data and the desired output to approximate a problem from a model-free perspective, which results in robust prediction performance against corruptions/imperfections in the wireless channel data.
2. It learns feature patterns that can be easily updated for new data and adapted to environmental changes, which leads to lower computational complexity in the long run.
3. ML-based solutions have significantly reduced runtimes due to their parallel processing capabilities,

which is not achievable in conventional optimization and signal processing algorithms.

### 5.1. System Model and Assumptions

Given the dynamic nature of high frequency changes in channel estimation, ML can prove to be a valuable tool. This section aims to provide an overview of the significant assumptions in the estimation model and present its two principal outcomes.

The present study involves a Multi-user MIMO-OFDM system with M subcarriers, wherein the BS communicates with K users possessing $N_{MS}$ antennas while itself having $N_{BS}$ antennas. In the downlink, K data symbols are first pre-coded by the BS using baseband precoders $F_{BB}$ at each subcarrier and subsequently transformed to the time domain via M-point inverse DFT. Following this, the BS adds a cyclic prefix and employs an analogue precoder $F_{RF}$ to form the transmitted signal. The transmitted signal then passes through a mm-wave channel, which is defined as $H_k[m]$, an $N_{MS} \times N_{BS}$ mm-wave channel matrix. Based on the assumptions and channel model [49], the received signal at the $k$th user before analog processing at subcarrier $m$ can be represented as $\tilde{y}_k[m] = \sqrt{\rho} H_k[m] x[m]$:

$$\tilde{y}_k[m] = \sqrt{\rho} H_k[m] F_{RF} F_{BB}[m] s[m] + n[m] \quad (16)$$

where $\rho$ represents the average received power and $n[m] \sim \mathcal{CN}(0, \sigma^2 I_{N_{MS}})$ is additive white Gaussian noise (AWGN) vector, and the received BB signal becomes:

$$\begin{aligned} \overline{y}_k[m] &= \sqrt{(\rho)} w_{RF,k}^H H_{(k)}[m] F_{(RF)} F_{(BB)}[m] s[m] \\ &+ w_{RF,k}^H n[m] \end{aligned} \quad (17)$$

where the analog combiner $w_{RF,k}$ has the constraint $\left[ w_{RF,k} w_{RF,k}^H \right]_{i,i} = 1$. To accurately identify data streams which are $s[m]$, estimation of channel matrix with FL is required.

In order to achieve this, the ChannelNet, a global neural network utilized for channel estimation, is located at the BS and trained on the local datasets of individual users. The local dataset for the $k$th user, denoted as $D_k$, comprises input-output pairs represented by $D_k^{(i)} = \left( X_k^{(i)}, Y_k^{(i)} \right)$, where $X$ denotes the received input/pilot signals and $Y_k^{(i)}$ represents the output/channel matrix. The size of the local dataset is given by $|Dk|$.

The ChannelNet establishes a non-linear relationship between input and output data, represented by $(X, Y)$, such that: $f(X|\theta) = Y$, where $\theta \in RP$ represents the learnable parameters.

The proposed FL approach consists of three stages: training data collection, where each user gathers their respective training dataset from the received pilot signals; model training that details the determination of input and output label data; and prediction of the estimated channel.

### 5.2. Learning Steps

Fig. 5 illustrates the communication interval for two consecutive data transmission blocks and presents the process of channel estimation at the beginning of each transmission block. Analytical channel estimation techniques, including compressed sensing [55], [56], angle-domain processing [53], and coordinated pilot-assignment [54], are used to process the received pilot signals for channel estimation. However, the analytical approach is only used in the training data collection stage, while ML/FL becomes more advantageous in the prediction stage in the long run [57]. Training data can be collected through offline datasets obtained from field measurements, but such data may not always reflect the mm-wave channel characteristics and imperfections.

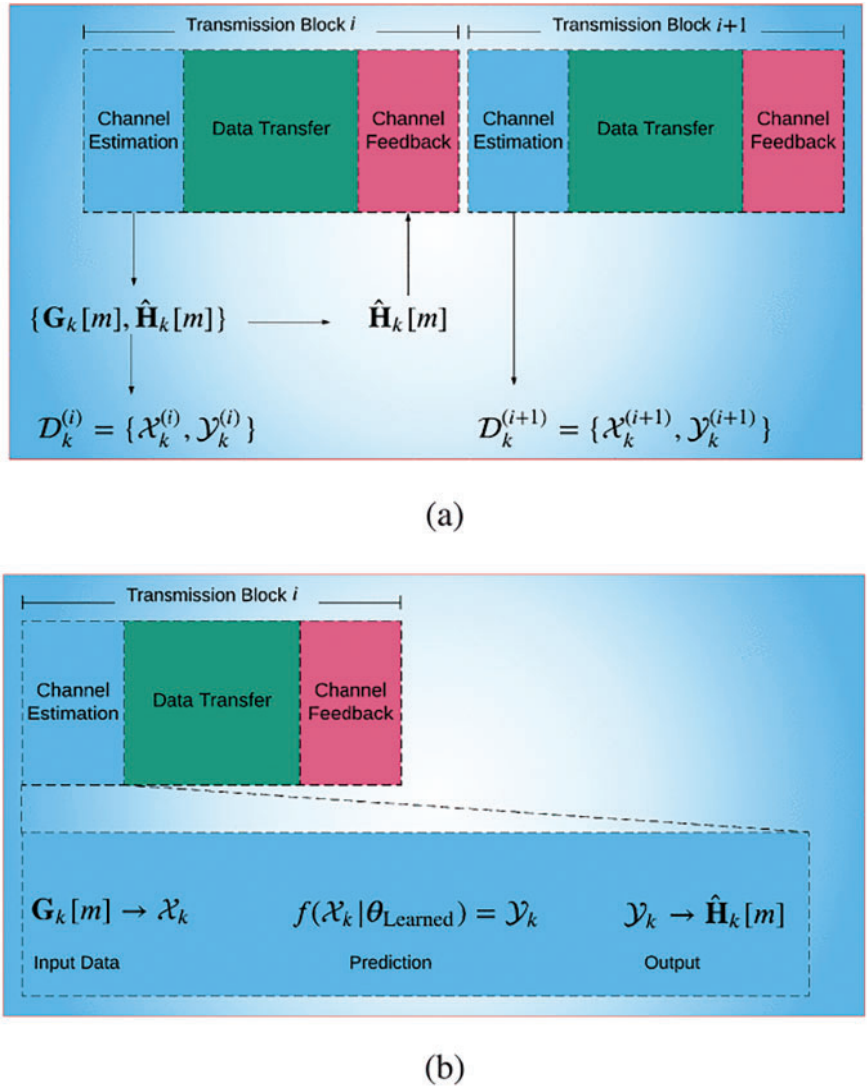

Fig. 5. Communication interval for consecutive transmission blocks and channel estimation process: (a) Training data collection and (b) channel estimation with the training model.

Evaluation of the performance of the proposed approach is done on datasets with both true and estimated channel data labels. After channel estimation, training data can be collected by storing received pilot data $G_k[m]$ and estimated channel data $\hat{H}_k[m]$. The collected local dataset $D_k$ allows to collect training data for different channel coherence times. This process is the first stage of the proposed FL-based channel estimation framework. Once the training data is collected, the global model is trained, and each user can estimate its own channel via the trained neural network by feeding $G_k[m]$ into the network and obtaining $\hat{H}_k[m]$, as shown in Fig. 5b.

Next Step would be FL-based model training. In FL, the local datasets are in user side and are not transmitted to BS side, so the training is done in user side as:

$$\underset{\theta}{minimize} \frac{1}{D_k} \sum_{i=1}^{D_k} \mathcal{L}\left(f\left(X_k^{(i)}|\theta_{t-1}\right), y_k^{(i)}\right) \quad (18)$$

$$Subject\ to: f\left(X_k^{(i)}|\theta_t\right) = y_k^{(i)}$$

$$\mathcal{L}_k(\theta) = \frac{1}{D_k} \sum_{i=1}^{D_k} \left\| f\left(X_k^{(i)}|\theta\right) - y_k^{(i)} \right\|_F^2 \quad (19)$$

In this research that is described here, for efficient solving of it, gradient descent is used, and problem is solved iteratively. Each use calculates the gradient $g_k(\theta_t) = \nabla \mathcal{L}_k(\theta_t)$ and then sends it to BS. Updating the learning parameter is done as followed:

$$\theta_{t+1} = \theta_t - \eta \frac{1}{K} = \sum_{k=1}^{K} g_k(\theta_t) \quad (20)$$

Transmitting gradients to the BS is a more energy-efficient approach compared to directly transmitting the model parameters, as in the FedAvg algorithm [58]. This is primarily because gradients contain only the model updates obtained from the GD algorithm, whereas model transmission includes previously known data from the previous iteration. Consequently, transmitting the model consumes a substantial amount of transmit power from all users.

As this modeling is done in wireless network, so effects of noise in transmission of gradients is inevitable, so this fact is considered in the model of the paper as well.

Based on aforementioned fact of noise effects, the actual model that is assumed and optimized is following ones which the effects of AWGN noise are added to it:

$$\underset{\theta}{minimize}\overline{\mathcal{L}}(\theta) = \frac{1}{K}\sum_{k=1}^{K}\tilde{\mathcal{L}}_k(\theta) \qquad (21)$$

$$Subject\ to: f\left(X_k^{(i)}|\theta\right) = y_k^{(i)}\tilde{\theta}_{t+1} = \tilde{\theta}_t - \eta\nabla\overline{\mathcal{L}}\left(\tilde{\theta}\right)$$

Because of noisy gradient transmission, $\overline{\mathcal{L}}(\theta)$ converges slower than $\mathcal{L}(\theta)$. While the paper provides a comprehensive account of the channel estimation steps for MIMO and RIS assisted scenarios, only a brief overview of the MIMO case will be presented here, and for more details can be referred to the original paper for a more explanation.

ChannelNet uses pilot signals during the initialization stage, assuming that the BS activates only one RF chain per step. The resulting beamformer vector is $\bar{f}_u[m] \in \mathbb{C}^{N_{BS}}$ and the pilot signals are $\bar{s}_u[m]$. At the receiver side, each user applies the beamformer vector $\bar{w}_v[m]$ for $M_{MS}$ times and activates the RF chain, processing the received pilots [59]. The total channel use in the acquisition process is $M_{BS}\left\lceil\frac{M_{MS}}{N_{RF}}\right\rceil$, resulting in $\overline{Y}_k[m]$ at the $k$th user given by (21):

$$\overline{Y}_k[m] = \overline{W}[m]H_k[m]\overline{F}[m]\overline{S}[m] + \tilde{N}_k[m] \qquad (22)$$

The beamformer matrices, $\overline{F}[m]$ and $\overline{W}[m]$, have dimensions of $N_{BS} \times M_{BS}$ and $N_{MS} \times M_{MS}$, respectively. Pilot signals are represented by $\overline{S}[m] = diag\{\bar{s}_1[m],\ldots,\bar{s}_{M_{BS}}[m]\}$. $\tilde{N}_k[m] = \overline{W}^H\overline{N}_k[m]$ is the effective noise matrix, where $\overline{N}_k[m] \sim \mathcal{N}(0,\sigma^2 I_{M_{MS}})$. Assuming $\overline{F}[m] = \overline{F}$ and $\overline{W}[m] = \overline{W}$, and $\overline{S}[m] = I_{M_{BS}}$ for all $m \in M$, the received signal (21) is expressed as:

$$\overline{Y}_k[m] = \overline{W}^H\overline{H}_k[m]\overline{F} + \tilde{N}_k[m] \qquad (23)$$

Using $\overline{Y}_k[m]$, the input of ChannelNet $G_k[m]$ as

$$G_k[m] = T_{MS}\overline{Y}_k[m]T_{BS} \qquad (24)$$

where

$$T_{MS} = \begin{cases} \overline{W}, & M_{MS} < N_{MS} \\ (\overline{W}\overline{W}^H)^{-1}\overline{W} & M_{MS} \geq N_{MS} \end{cases} \qquad (25)$$

and

$$T_{BS} = \begin{cases} \overline{F}^H, & M_{BS} < N_{BS} \\ \overline{F}^H(\overline{F}\overline{F}^H)^{-1} & M_{BS} \geq N_{BS} \end{cases} \qquad (26)$$

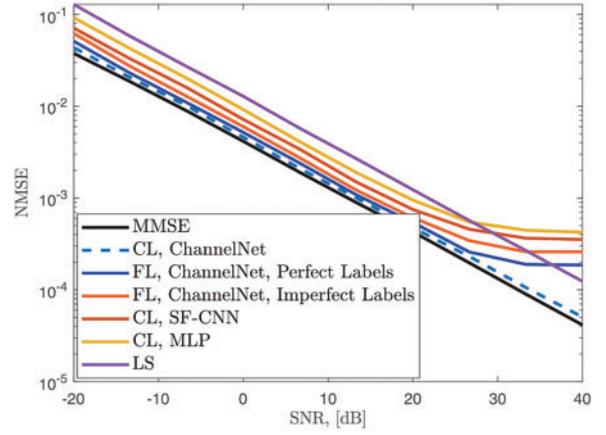

Fig. 6. Channel estimation for different algorithms in MIMO.

The output of ChannelNet system is given by a $2N_{BS}N_{MS} \times 1$ real-valued vector as:

$$y_k = [vec\{Re\{H_k[m]\}\}^T, vec\{Im\{H_k[m]\}\}^T]^T \qquad (27)$$

### 5.3. Neural Network Architecture

The proposed network employs a 10-layer convolutional neural network, which takes $y_k$ as input and outputs a channel matrix. The specifics of the layers and other details are not discussed here.

One of the most important parameters in FL after privacy is transmission overhead, which refers to the data transmitted during model training. Using this definition, the overheads of Federated Learning (FL) and Centralized Learning (CL) are compared. In CL, the number of symbols identifies it in the uplink, while in FL, the gradient of the learning parameter in the uplink and the value of the learning parameter in the downlink specify it. Therefore, it is evident that the overhead of CL is significantly higher than that of FL.

### 5.4. Simulation Results

The performance of the proposed CNN architecture in channel estimation is evaluated using both perfect and imperfect labels. It can be observed a slight decrease in performance in the case of imperfect labels compared to perfect CSI. However, the accuracy of the channel estimation algorithm used during the collection of the training dataset strongly affects the performance of the imperfect label scenario.

Fig. 6 represents the channel estimation based on different algorithms, and as it is clear, CL follows the performance of NMSE closely. FL is working well when SNR < 25 dB and its performance max out after that as it has no access to entire datasets like CL [60].

Fig. 7a presents the training performance and the channel estimation NMSE (Fig. 7b) of the proposed FL approach for channel estimation with varying numbers of users. The total dataset size D is fixed by selecting $G = 20 \cdot 8/K$. As K increases, the training performance improves and approaches the performance of CL, as the model updates superposed at the BS become more robust against noise. Conversely, as K decreases, the corruptions in the model aggregation increase due to the diversity in the training dataset [60].

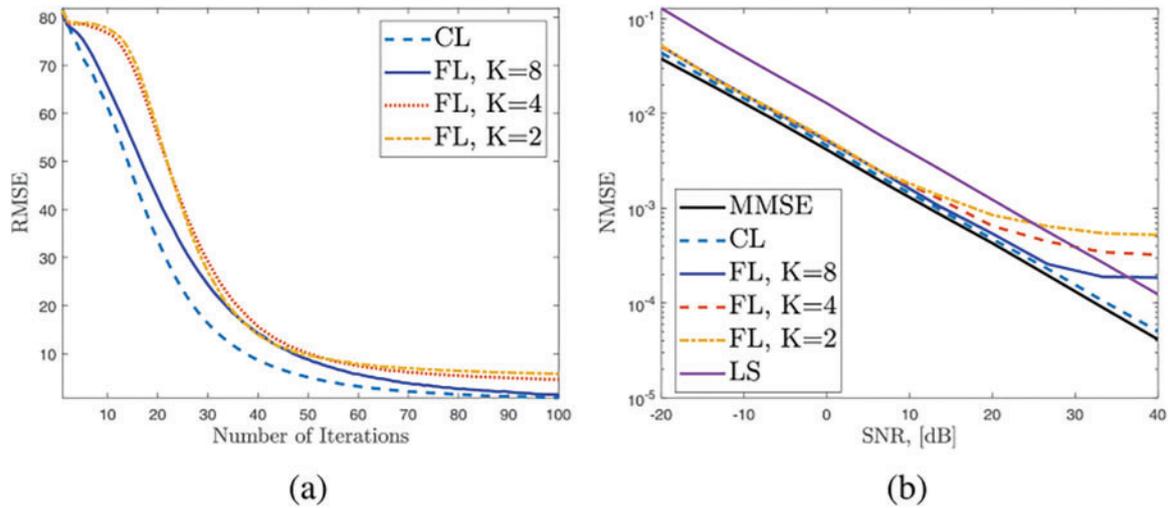

Fig. 7. Training performance and channel estimation results of the proposed FL approach for varying numbers of users in a MIMO system: (a) Training performance and (b) Channel estimation with respect to K in MIMO.

## 6. Conclusion

In this report, we have provided an overview of machine learning and its significance. We then discussed federated learning, a distributed learning scheme that is particularly beneficial in wireless communication systems and can protect user privacy. Additionally, we presented an example of using federated learning to address the problem of channel estimation in MIMO channels.

Federated learning is gaining popularity in various applications, and as new generations of wireless systems emerge, there is a growing inclination towards adopting this scheme. Moreover, there are numerous challenges and research directions that need to be explored to facilitate the widespread use of federated learning in wireless networks.

To enhance the efficacy of federated learning in wireless networks, further investigation is required in areas such as model optimization, resource allocation, network architecture, and security. In conclusion, we believe that federated learning has enormous potential to revolutionize the wireless industry, and we look forward to seeing more advancements in this area.

## Conflict of Interest

The authors declare that they do not have any conflict of interest.

## References

[1] Gündüz D, Kurka DB, Jankowski M, Amiri MM, Ozfatura E, Sreekumar S. Communicate to learn at the edge. *IEEE Commun Mag*. 2020;58(12):14–9.

[2] Park J, Samarakoon S, Bennis M, Debbah M. Wireless network intelligence at the edge. *Proc IEEE*. 2019;107(11):2204–39.

[3] Amiri MM, Gündüz D. Machine learning at the wireless edge: distributed stochastic gradient descent over-the-air. *IEEE Trans Signal Process*. 2020;68:2155–69.

[4] Amiri MM, Gündüz D. Federated learning over wireless fading channels. *IEEE Trans Wirel Commun*. 2020;19(5):3546–57.

[5] Zhang W, Gupta S, Lian X, Liu J. Staleness-aware async-SGD for distributed deep learning. *Proceeding 25th International Joint Conferences on Artificial Intelligence (IJCAI)*, pp. 2350–6, 2016.

[6] Lian X, Huang Y, Li Y, Liu J. Asynchronous parallel stochastic gradient for nonconvex optimization. In *Advances in Neural Information Processing Systems*. vol. 28, Cortes C, *et al.*, Eds. Curran Associates, Inc., 2015, pp. 2737–45.

[7] Zheng S, Meng Q, Wang T, Chen W, Yu N, Ma ZM, *et al*. Asynchronous stochastic gradient descent with delay compensation. In *International Conference on Machine Learning*. PMLR, 2017 Jul, pp. 4120–9.

[8] Lian X, Zhang C, Zhang H, Hsieh CJ, Zhang W, Liu J. Can decentralized algorithms outperform centralized algorithms? A case study for decentralized parallel stochastic gradient descent. *Adv Neural Inform Process Syst*. 2017;30:5330–40.

[9] Tang H, Lian X, Yan M, Zhang C, Liu J. d2: decentralized training over decentralized data. In *Proceedings of the 35th International Conference on Machine Learning*. Dy J, Krause A, Eds. 2018, pp. 4848–56.

[10] Eghlidi NF, Jaggi M. *Sparse Communication for Training Deep Networks*. arXiv preprint, arXiv 2009.09271, 2020.

[11] Wangni J, Wang J, Liu J, Zhang T. Gradient sparsification for communication-efficient distributed optimization. In *Advances in Neural Information Processing Systems*. vol. 31, Bengio S, *et al.*, Eds. Curran Associates, Inc., 2018, pp. 1305–15.

[12] Jiang P, Agrawal G. A linear speedup analysis of distributed deep learning with sparse and quantized communication. In *Advances in Neural Information Processing Systems*. vol. 31, Bengio S, *et al.*, Eds. Curran Associates, Inc., 2018, pp. 2529–40.

[13] Aji AF, Heafield K. Sparse communication for distributed gradient descent. *Proceeding Conference on Empirical Methods in Natural Language Processing*, pp. 440–5, 2017.

[14] Alistarh D, Hoefler T, Johansson M, Konstantinov N, Khirirat S, Renggli C. The convergence of sparsified gradient methods. In *Advances in Neural Information Processing Systems*. vol. 31, Bengio S, *et al.*, Eds. Curran Associates, Inc., 2018, pp. 5976–86.

[15] Stich SU, Cordonnier JB, Jaggi M. Sparsified SGD with memory. In *Advances in Neural Information Processing Systems*. vol. 31, Bengio S, *et al.*, Eds. Curran Associates, Inc., 2018, pp. 4448–59.

[16] Simonyan K, Zisserman A. Very deep convolutional networks for large-scale image recognition. *International Conference on Learning Representations*, 2015.

[17] Karimireddy SP, Rebjock Q, Stich S, Jaggi M. Error feedback fixes signsgd and other gradient compression schemes. In *International Conference on Machine Learning*. PMLR, 2019 May, pp. 3252–61.

[18] Yu H, Yang S, Zhu S. Parallel restarted SGD with faster convergence and less communication: demystifying why model averaging works for deep learning. arXiv preprint, arXiv 1807.06629, 2018.

[19] Haddadpour F, Kamani MM, Mahdavi M, Cadambe V. Local SGD with periodic averaging: tighter analysis and adaptive synchronization. In *Advances in Neural Information Processing Systems*. vol. 32, Curran Associates Inc., 2019, pp. 11082–94.

[20] Haddadpour F, Kamani MM, Mahdavi M, Cadambe V. Ensemble distillation for robust model fusion in federated learning. arXiv preprint, arXiv 2006.07242, 2020.

[21] Jeong E, Oh S, Kim H, Park J, Bennis M, Kim SL. Communication-efficient on-device machine learning: federated distillation and

augmentation under non-iid private data. arXiv preprint, arXiv 1811.11479, 2018.
[22] Polyak BT. Some methods of speeding up the convergence of iteration methods. *USSR Comput Math Math Phys*. 1964 Dec;4(5):1–17.
[23] Li W, Milletarì F, Xu D, Rieke N, Hancox J, Zhu W, *et al.* Privacy-preserving federated brain tumour segmentation. In *Machine Learning in Medical Imaging*. Suk H-I, *et al.*, Eds. Springer International Publishing, 2019, pp. 133–41.
[24] Li W, Milletarì F, Xu D, Rieke N, Hancox J, Zhu W, *et al.* The future of digital health with federated learning. arXiv preprint, arXiv 2003.08119, 2020.
[25] Xia W, Quek TQ, Guo K, Wen W, Yang HH, Zhu H. Multi-armed bandit-based client scheduling for federated learning. *IEEE Transactions on Wireless Communications*. 2020;19(11):7108–23.
[26] Yang HH, Arafa A, Quek TQ, Poor HV. Age-based scheduling policy for federated learning in mobile edge networks. *IEEE International Conference on Acoustics, Speech and Signal Processing (ICASSP)*, pp. 8743–7, 2020.
[27] Yang HH, *et al.* Scheduling policies for federated learning in wireless networks. *IEEE Trans Commun*. 2020;68(1):317–33.
[28] Shi W, Zhou S, Niu Z. Device scheduling with fast convergence for wireless federated learning. *IEEE International Conference on Communications (ICC)*, pp. 1–6, 2020.
[29] Konečný J, McMahan B, Ramage D. Federated optimization: distributed optimization beyond the datacenter. arXiv:1511.03575, 2015 Nov. Available from: https://arxiv.org/abs/1511.03575.
[30] Chen T, Giannakis G, Sun T, Yin W. LAG: lazily aggregated gradient for communication-efficient distributed learning. *Proceedings of the Advances in Neural Information Processing System*, pp. 5050–60, 2018.
[31] Aji AF, Heafield K. Sparse communication for distributed gradient descent. *Conference Empirical Methods Natural Language Process (EMNLP)*, pp. 440–5, 2017 Sep.
[32] Lin Y, Han S, Mao H, Wang Y, Dally WJ. Deep gradient compression: reducing the communication bandwidth for distributed training. *Proceeding International Conference Learning Represent (ICLR)*, pp. 1–14, 2018 May.
[33] Wang X, Han Y, Wang C, Zhao Q, Chen X, Chen M. In-Edge AI: intelligentizing mobile edge computing caching and communication by federated learning. arXiv:1809.07857, 2018 Sep. Available from: https://arxiv.org/abs/1809.07857.
[34] Wang S, Tuor T, Salonidis T, Leung KK, Makaya C, He T, *et al.* When edge meets learning: adaptive control for resource-constrained distributed machine learning. *Proceeding of the IEEE INFOCOM*, pp. 63–71, 2018 Apr.
[35] Wang S, Tuor T, Salonidis T, Leung KK, Makaya C, He T, *et al.* Adaptive federated learning in resource constrained edge computing systems. *IEEE J Sel Areas Commun*. 2019 Jun;37(3):1205–21.
[36] Nishio T, Yonetani R. Client selection for federated learning with heterogeneous resources in mobile edge. arXiv:1804.08333, 2018 Apr. Available from: https://arxiv.org/abs/1804.08333.
[37] Zhu G, Wang Y, Huang K. Broadband analog aggregation for low-latency federated edge learning (extended version). arXiv:1812.11494, 2018 Dec. Available from: https://arxiv.org/abs/1812.11494.
[38] Yang K, Jiang T, Shi Y, Ding Z. Federated learning via over-the-air computation. arXiv:1812.11750, 2018 Dec. Available from: https://arxiv.org/abs/1812.11750.
[39] Ha S, Zhang J, Simeone O, Kang J. Coded federated computing in wireless networks with straggling devices and imperfect CSI. arXiv:1901.05239, 2019 Jan. Available from: https://arxiv.org/abs/1901.05239.
[40] Lan G, Lee S, Zhou Y. Communication-efficient algorithms for decentralized and stochastic optimization. *Math Program*. 2020;180(1):237–84.
[41] Nishio T, Yonetani R. Client selection for federated learning with heterogeneous resources in mobile edge. arXiv:1804.08333, 2018 Apr. Available from: https://arxiv.org/abs/1804.08333.
[42] Jiang Y, Wang S, Valls V, Ko BJ, Lee WH, Leung KK, *et al*. Model pruning enables efficient federated learning on edge devices. arXiv preprint, arXiv:1909.12326, 2019.
[43] Goodfellow I, Bengio Y, Courville A. *Deep Learning*. MIT Press; 2016.
[44] McMahan B, Moore E, Ramage D, Hampson S, Arcas BA. Communication-efficient learning of deep networks from decentralized data. In *Artificial Intelligence and Statistics*. PMLR, 2017, Apr, pp. 1272–82.
[45] Tuor T, Wang S, Leung KK, Chan K. Distributed machine learning in coalition environments: overview of techniques. *21st International Conference on Information Fusion (FUSION)*, pp. 814–21, 2018.
[46] Li T, Sahu AK, Zaheer M, Sanjabi M, Talwalkar A, Smith V. Federated optimization in heterogeneous networks. *Conference on Machine Learning and Systems (MLSys)*, 2020.
[47] Yurochkin M, Agarwal M, Ghosh S, Greenewald K, Hoang N, Khazaeni Y. Bayesian nonparametric federated learning of neural networks. *Proc Mach Learn Res (PMLR)*. 2019 Jun;97:7252–61.
[48] Björnson E, Van der Perre L, Buzzi S, Larsson EG. Massive MIMO in sub-6 GHz and mmWave: physical practical and use-case differences. *IEEE Wireless Commun*. 2019 Apr;26(2):100–8.
[49] Alkhateeb A, Heath RW. Frequency selective hybrid precoding for limited feedback millimeter wave systems. *IEEE Trans Commun*. 2016 May;64(5):1801–18.
[50] Sohrabi F, Yu W. Hybrid analog and digital beamforming for mmWave OFDM large-scale antenna arrays. *IEEE J Sel Areas Commun*. 2017 Jul;35(7):1432–43.
[51] Huang C, Zappone A, Alexandropoulos GC, Debbah M, Yuen C. Reconfigurable intelligent surfaces for energy efficiency in wireless communication. *IEEE Trans Wireless Commun*. 2019 Aug;18(8):4157–70.
[52] Taha A, Alrabeiah M, Alkhateeb A. Enabling large intelligent surfaces with compressive sensing and deep learning. *arXiv:1904.10136*, 2019.
[53] Fan D, Gao F, Liu Y, Deng Y, Wang G, Zhong Z, *et al.* Angle domain channel estimation in hybrid millimeter wave massive MIMO systems. *IEEE Trans Wireless Commun*. 2018 Dec;17(12):8165–79.
[54] Yin H, Gesbert D, Filippou M, Liu Y. A coordinated approach to channel estimation in large-scale multiple-antenna systems. *IEEE J Sel Areas Commun*. 2013 Mar;31(2):264–73.
[55] Bajwa WU, Haupt J, Raz G, Nowak R. Compressed channel sensing. *Proceeding 42nd Annual Conference Information Science System*, pp. 5–10, 2008 Mar.
[56] Marzi Z, Ramasamy D, Madhow U. Compressive channel estimation and tracking for large arrays in mm-wave picocells. *IEEE J Sel Topics Signal Process*. 2016 Apr;10(3):514–27.
[57] Elbir AM, Mishra KV, Shankar MRB, Ottersten B. A family of deep learning architectures for channel estimation and hybrid beamforming in multi-carrier mm-wave massive MIMO. *arXiv:1912.10036*, 2019.
[58] McMahan HB, Moore E, Ramage D, Hampson S, Agüera y Arcas B. Communication-efficient learning of deep networks from decentralized data. *arXiv:1602.05629*, 2016.
[59] Dong P, Zhang H, Li GY, Gaspar I, NaderiAlizadeh N. Deep CNN-based channel estimation for mmWave massive MIMO systems. *IEEE J Sel Topics Signal Process*. 2019 Sep;13(5):989–1000.
[60] Elbir AM, Coleri S. Federated learning for channel estimation in conventional and RIS-assisted massive MIMO. *IEEE Trans Wirel Commun*. 2022 Jun;21(6):4255–68. doi: 10.1109/TWC.2021.3128392.
[61] Yang HH, Liu Z, Quek TQS, Poor HV. Scheduling policies for federated learning in wireless networks. *IEEE Trans Commun*. 2020 Jan;68(1):317–33. doi: 10.1109/TCOMM.2019.2944169.